# Korean *aegyo* speech shows systematic F1 increase to signal childlike qualities

Ji-eun Kim[1a] and Volker Dellwo[2]

*[1] Department of Korean Language and Literature, Duksung Women's University, Seoul, 01369, South Korea*

*[2] Department of Computational Linguistics, University of Zurich, Zürich, 8050, Switzerland*

*smart173@duksung.ac.kr, volker.dellwo@uzh.ch*

Korean *aegyo* is a socially recognized childlike speaking style used predominantly in romantic interactions among adults. This study examined vowel space modification in *aegyo* by analyzing formant frequencies from twelve Seoul Korean speakers who produced identical scripts in *aegyo* and non-*aegyo* styles. Results show that *aegyo* speech features a significant increase in F1 values across vowels and selective fronting of front vowels, leading to vowel space expansion but mainly a shift to higher F1. These findings suggest that adult speakers stylize childlike speech by imitating the shorter vocal tract of children, mainly through global vowel lowering and partial fronting.

---

[1a] Author to whom correspondence should be addressed.

1. **Introduction**

Often described as a babyish or childlike way of speaking, Korean *aegyo* is used predominantly in romantic adult communication to convey "a childlike charm and infantilized cuteness" (Puzar and Hong, 2018). Korean speakers have a high awareness of *aegyo*; they can evaluate whether *aegyo* is well or poorly performed and consider it as a conventionalized way of speaking rather than an idiosyncratic individual habit (Strong, 2012; Kim and Dellwo, 2025). In this sense, *aegyo* is a useful case for examining how a socially recognizable childlike persona is phonetically constructed in adult speech.

Previous studies on *aegyo* have shown that it is associated with higher mean pitch (Strong, 2012; Kim and Dellwo, 2025), obstruent fortition (Crosby and Dalola, 2023a), nasalance (Crosby and Dalola, 2023b), slower speech rate, and larger temporal and pitch variability (Kim and Dellwo, 2025). However, the vowel space of *aegyo* has not yet been examined. This is an important gap, because vowel space is one of the phonetic dimensions that speakers can systematically manipulate to signal different speaking styles. For example, a large body of research on Infant-Directed Speech (IDS) has shown that vowel triangle area shaped by corner vowels such as /i, a, u/ are realized in more peripheral positions in IDS than in Adult-Directed Speech (ADS) (Kuhl et al., 1997; Burnham et al., 2003; Uther et al., 2007). Likewise, Xu et al. (2023) reported that vowel triangle expansion systematically varies across speaking styles, being strongest in IDS, weaker in parrot-directed speech, and absent in dog-directed speech.

The vowel space of *aegyo* is also relevant in line with previous research on children's speech. As children have a shorter vocal tract, they typically exhibit higher formant frequencies, especially in F1, and these anatomical differences directly influence the scaling of the vowel space (Vorperian et al, 2005; Yoon and Ha, 2022). Meanwhile, studies of children's vowel space have shown conflicting results: some report larger spaces than adults (Pettinato et al., 2016), whereas others report smaller spaces despite similar overall shape (Chung et al., 2012). This difference has been attributed in part to whether normalization for vocal tract length is applied, since anatomical differences between children and adults directly affect formant scaling (Yoon and Ha, 2022).

In this study, we were interested in the contribution of vowel space to how speakers construct a child-like speaking style. To this end, we examined whether, and if so how, Korean *aegyo* speakers manipulate their vowel space, using both traditional corner-vowel triangles shaped by /i, a, u/ and a vowel space defined by the full monophthong inventory. We included these two vowel spaces because the triangle is directly comparable to much of the previous research on children's speech (e.g. Vorperian and Kent, 2007; Chung et al., 2012; Pettinato et al., 2016) whereas the full-vowel space allows us to test whether the same pattern as the corner-vowel space applies across the vowel system as a whole. Since *aegyo* is often perceived as a childlike speaking style, our data will be useful for interpreting how children's vowel space is perceptually represented and enregistered (Agha, 2005) by adult speakers in Korea. Both *aegyo* and non-*aegyo* speech in our data is produced by the same adult speakers.

## 2. Methods

### 2.1 Speakers and apparatus

Twelve Seoul Korean speakers (6 females; 6 males) between the ages of 25 and 31 took part in the production experiment. Recordings were conducted in a sound-treated booth in the Phonetics Lab at Seoul National University using a Tascam DR-100MKIII recorder and a SHURE 10A head-worn microphone (sampling rate = 44.1kHz, quantization level = 16 bit). All participants were free from any speech or hearing disorders.

### 2.2 Reading materials and procedures

Reading materials comprised 31 sentences. Participants first read two sets of Korean numerals, from one to ten to familiarize themselves with the recording environment. The following main task consisted of 29 sentences presented as a narrative script for a commercial. Participants read the identical script in both styles consecutively. This design allowed for direct comparison of vowel spaces while controlling for vowel categories and their phonological conditions.

To elicit the target styles, we explicitly used the metadiscursive label *aegyo* in the instructions: "Please read the script using *aegyo* fully" for the *aegyo* speech and "Please read the script without using any *aegyo*" for the non-*aegyo* speech. All participants reported that they clearly understood the term *aegyo* and how to perform it, and none expressed uncertainty or confusion regarding the instructions.

*2.3    Annotation*

Segmental intervals were automatically aligned (Yoon and Kang, 2013) and were manually corrected by the first author. To facilitate the extraction of vowel tokens, an additional tier was created to indicate consonant and vowel segments. Based on these markings, vowel intervals were identified from the aligned segmental tier and organized into a separate tier for acoustic measurements used in the vowel space analysis.

Each segment was annotated based on the standard Seoul Korean pronunciation (National Institute of Korean Language, 2026) corresponding to its orthographic form in the script, and all analyses were conducted on the basis of these phonemic labels. This study adopts an eight-vowel system /a, e, ɛ, i, o, u, ʌ, ɯ/ for the annotation and analysis of Seoul Korean. In terms of vowel height, this eight-vowel system comprises three high vowels (/i, u, ɯ/), two mid vowels (/e, ʌ, o/), and two low vowels /ɛ, a/. In terms of backness, this system is characterized by three front vowels (/i, e, ɛ/) and five back vowels (/a, o, u, ʌ, ɯ/).

Meanwhile, the complete vowel inventory of Standard Seoul Korean is described as having ten monophthongs /a, e, ɛ, i, o, u, ʌ, ɯ, ø, y/ (National Institute of Korean Language, 2026). The front rounded vowels /ø/ and /y/, however, are undergoing diphthongization, particularly among younger speakers (Heo, 2013), with lexical variation. In our data, none were produced as monophthongs, and they were therefore omitted from the analysis. It should also be noted that the merger of /e/ and /ɛ/ is ongoing in Seoul Korean (Eychenne and Jang, 2018). In addition, recent studies have reported increasing similarity between /o/ and /u/ in the F1-F2 space, associated with lowering of F1 for /o/ (Kang and Kong, 2016; Yoon and Ha, 2022). Consequently, some speakers may show minimal F1-F2 separation for /e/-/ɛ/ and /o/-/u/, even though these vowels were labeled and analyzed separately according to their underlying phonemic categories.

*2.4    Acoustic analysis*

For each utterance, a Formant object was created in Praat (Boersma and Weenink, 2026) to obtain formant frequency contours. We used a standard LPC-based Burg algorithm with a 25 ms window length, a 10 ms time step, pre-emphasis from 50 Hz, a maximum of five formants, and a formant ceiling of 5,500 Hz. For each vowel token, mean F1 and F2 were calculated over the central 40% of its duration to reduce coarticulatory effects. For vowel space plotting, F1 and F2 values were z-score normalized by speaker to control for between-speaker variability.

Vowel space measures were computed at the utterance level (i.e., speaker * item * style) using the vowel-category means within that utterance. For the corner-vowel space, mean F1 and F2 were calculated for /i/, /a/, and /u/, and the vowel space area was defined as the area of the triangle formed by these three mean points. For the all-vowel space, mean F1 and F2 were calculated for each of the eight monophthongs (/a, e, ɛ, i, o, u, ʌ, ɯ/), and the vowel space area was defined as the area of the convex hull covering these mean points. Centroids were computed separately for the corner-vowel set and the all-vowel set as the arithmetic mean of the corresponding vowel-category mean coordinates: centroid F1 was the mean of the included vowels' mean F1 values, and centroid F2 was the mean of the included vowels' mean F2 values.

*2.5     Statistical analysis*

Linear mixed-effects models were fitted using lme4 package (Bates et al., 2015) in R (R Core Team, 2021), with p-values obtained via lmerTest package (Kuznetsova et al., 2017). For each dependent variable, a maximal model was initially specified, including by-speaker random slopes for style and random intercepts for speaker and item. The dependent variables were vowel space area, centroid F1, and centroid F2 at both the corner-vowel and all-vowel levels, as well as vowel-level F1 and F2 values. When models failed to converge or produced singular fits, the random-effects structure was reduced by removing random slopes. The final models for F2 centroids of corner vowels, vowel space areas and F1 centroids only included random intercepts for speaker and item, whereas the rest of the models retained the random slopes. Vowel-level formant analyses included fixed effects of style, vowel, and their interaction. Sum coding was applied to both style and vowel for interpreting the global main effects independent of baseline levels. Pairwise comparisons were conducted using estimated marginal means with Holm correction. Statistical significance was evaluated at $p < .05$.

## 3. Results

### 3.1 Style effects on vowel space areas and centroids

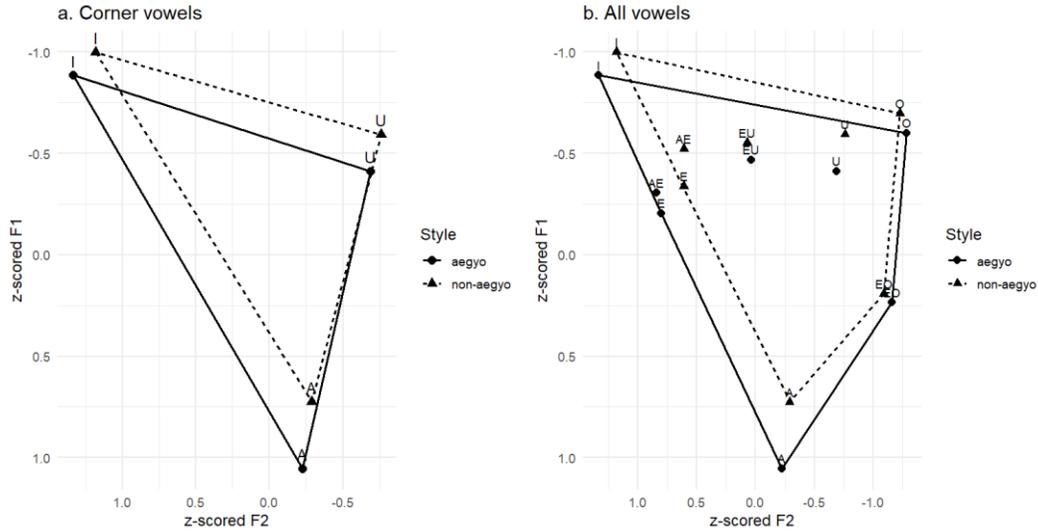

Fig. 1. Z-scored vowel spaces in the *aegyo* and non-*aegyo* conditions, defined (a) by the corner-vowel triangle /i, a, u/ and (b) by the convex hull of all monophthongs /a, e, ɛ, i, o, u, ʌ, ɯ/.

Figure 1a displays the corner vowel space in *aegyo* and non-*aegyo* speech. The plot is based on F1 and F2 values z-scored by speaker, after excluding outliers using IQR filtering within each speaker * style * vowel group and retaining only groups with at least five tokens. For each style, mean F1 and F2 were calculated for I /i/, A /a/ and U /u/, and the three mean points were connected to form the vowel space area. Visually, the *aegyo* area is slightly larger (mainly because of /I/ shifting forward and downwards) and shows a downward shift with a leftward shift (corner vowels: 1.59 vs. 1.38; all vowels: 2.67 vs. 2.18 for *aegyo* and non-*aegyo*, respectively). The same pattern is observed in the raw formant values: the corner-vowel triangle area is 144,493 Hz² in *aegyo* and 126,777 Hz² in non-*aegyo*, and the convex-hull area for all vowels is 240,422 Hz² in *aegyo* and 193,968 Hz² in non-*aegyo*.

To evaluate these patterns statistically, a linear mixed-effects model was fitted to the corner-vowel space area with style as a fixed effect and random intercepts for speaker and item. The same random-effects structure was used for the F1 and F2 centroid models, as maximal models including by-speaker random slopes for style did not converge. The models show that *aegyo* is associated with both expansion and lowering and fronting of the corner vowel space. The vowel space area defined by the corner vowels /i, a, u/ was significantly larger in the *aegyo* condition (b = −19,400.56 Hz², SE = 8,271.70, t = −2.35, p = .0197). Centroid analyses showed parallel positional shifts. The F1 centroid showed a significant style effect (b = −43.27 Hz, SE = 5.14, t = −8.43, p < .001), indicating a downward shift in the *aegyo* condition (i.e., increased F1 values). The F2 centroid also showed a significant style effect (b = −56.38 Hz, SE = 14.34, t = −3.93, p < .001), indicating a forward shift in the *aegyo* condition (i.e., increased F2 values).

Figure 1b displays the vowel space defined by the convex hull of all vowels in *aegyo* and non-*aegyo* speech. For each style, mean F1 and F2 were calculated for A /a/, E /e/, AE /ɛ/, I /i/, O /o/, U /u/, EO /ʌ/ and EU /ɯ/, and the eight mean points were connected to form the vowel space area. Visually, the *aegyo* area is larger and shows a downward shift with a leftward shift. It should also be noted that the relatively high positions of /ɛ/ over /e/ and /o/ over /u/ in Figure 1b are consistent with previously reported sound changes in Seoul Korean, namely the ongoing /e/-/ɛ/ merger (Eychenne and Jang, 2018) and the increasing /o/-/u/ similarity associated with raising of /o/ (Kang and Kong, 2016).

Linear mixed-effects models were fitted to the vowel space defined by the all-vowel space with style as a fixed effect and random intercepts for speaker and item for both the area and F1 centroid measures. Only for the F2 centroid, the final model retained by-speaker random slopes for style. For the vowel space calculated using the convex hull of all monophthongs /a, e, ɛ, i, o, u, ʌ, ɯ/, *aegyo* is associated with significant expansion and a downward shift along the F1 dimension, but no significant shift along the F2 dimension. The overall vowel space area was significantly larger in the *aegyo* condition (b = −38,978.73 Hz², SE = 5,822.42, t = −6.70, p < .001). Centroid measures showed a significant effect in the F1 dimension (b = −26.68 Hz, SE = 3.11, t = −8.58, p < .001), indicating a global vowel lowering of vowel height in the *aegyo* condition (i.e., increased F1 values). In contrast, the F2 centroid did not exhibit a significant style effect (b = −24.80 Hz, SE = 20.36, t = −1.22, p = .249), indicating no significant front-back shift across the full vowel inventory.

## 3.2 Style effects on formant values

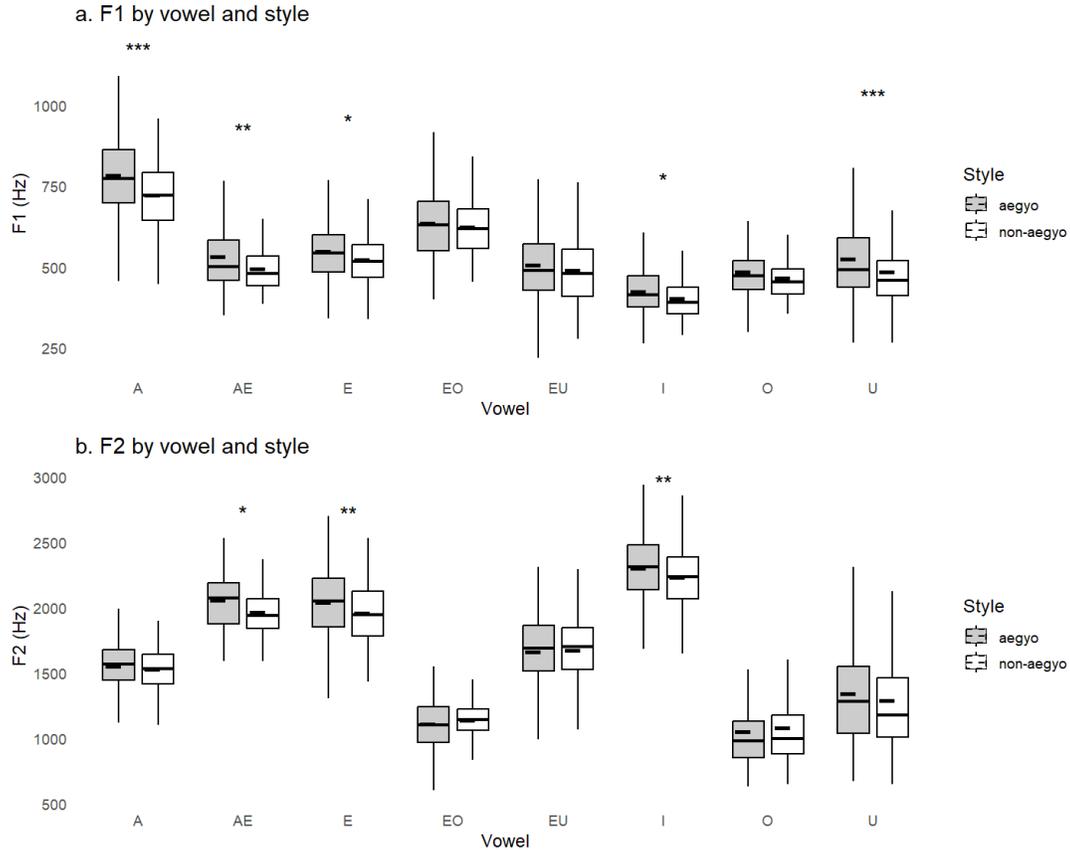

Fig. 2. Boxplots of raw (a) first and (b) second formant values (Hz) by vowel and style. In the boxplots, boxes indicate the interquartile range, solid lines indicate the median, and dashed lines indicate the mean.

Figure 2a displays the distribution of F1 values (Hz) across vowels in the *aegyo* and non-*aegyo* styles. Visually, F1 values are generally higher in the *aegyo* condition, with larger differences for certain vowels.

A linear mixed-effects model was fitted to F1 values with style and vowel as fixed effects and by-speaker random slopes for style, along with random intercepts for item. The model showed a significant global effect of style (b = −27.73 Hz, SE = 6.92, t = -4.01, p = .0015), indicating that vowels in the *aegyo* condition were produced with significantly increased F1 values across the vowel inventory. Pairwise comparisons showed that this lowering effect was significant only for a subset of vowels. The largest effect was observed for the /a/ (b = −59.7 Hz, SE = 8.73, z = −6.85, p < .0001). Significant lowering was also found for /ɛ/ (b = −36.6 Hz, SE = 14.00, z = −2.61, p = .009), /e/ (b = −23.1 Hz, SE = 10.50, z = −2.21, p = .027) and /i/ (b = −21.9 Hz, SE = 8.92, z = −2.46, p = .014), and /u/ (b = −35.8 Hz, SE = 10.60, z = −3.39, p = .0007). Effects for /o/, /ʌ/ and /ɯ/ did not show significant differences. Asterisks in Figure 4 indicate vowels that show significant style differences based on the pairwise comparisons (p < .05*, p < .01**, p <.001***).

Figure 2b displays the distribution of F2 values (Hz) across vowels in the *aegyo* and non-*aegyo* styles. Visually, F2 values are higher for front vowels I /i/, E /e/, AE /ɛ/ in the *aegyo* condition.

A linear mixed-effects model was fitted to F2 values with style and vowel as fixed effects and by-speaker random slopes for style, along with random intercepts for item. The model did not show a significant global effect of style (p = .063), indicating no reliable overall shift in vowel backness across the inventory. However, vowel-specific contrasts revealed selective fronting effects. Pairwise comparisons showed significant increases in F2 in the *aegyo* condition for the front vowels /i/ (b = −73.57 Hz, SE = 23.5, z = −3.13, p = .0017), /ɛ/ (b = −98.45 Hz, SE = 40.6, z = −2.43, p = .015), and /e/ (b = −75.45 Hz, SE = 28.9, z = −2.61, p = .009). Effects for back vowels /ɯ, ʌ, a, u, o/ did not show significant differences.

## 4. Discussion

This study examined vowel space in Korean *aegyo*, a socially enregistered childlike speaking style in Korea (Strong, 2012; Puzar and Hong, 2018). The results show that the most consistent acoustic effect of *aegyo* is a systematic increase in F1. Although *aegyo* is associated with a larger vowel space, our data shows that the increase in F1 is the more dominant phenomenon. The expansion is primarily driven by global vowel lowering (i.e., increased F1 values) and selective fronting of front vowels (i.e., increased F2 values), rather than by symmetrical enhancement of all vowel contrasts. This suggests that the changes from non-*aegyo* to *aegyo* may reflect an attempt to approximate some acoustic characteristics associated with child-like speech (see Fig. 1, 2).

The pattern of expansion also differs depending on how vowel space is defined. For the corner vowels (/i, a, u/), both F1 and F2 centroids showed significant shifts, suggesting expansion with lowering and fronting. However, when the full vowel inventory was considered, the effect on F2 was no longer significant. This indicates that fronting is limited to a subset of vowels, rather than the entire inventory. Therefore, analyses limited to corner vowels may therefore overestimate contrast enhancement in *aegyo* speech.

Taken together, these findings suggest that vowel space expansion in *aegyo* is best understood not as uniform dispersion of vowel categories, but as a shifted configuration of the vowel space along the F1 axis. The most plausible interpretation of the robust F1 increase in *aegyo* is that speakers approximate some acoustic consequences of a shorter vocal tract in order to sound more childlike. In particular, raising the larynx would shorten the pharyngeal cavity and could contribute to the observed F1 increase. This interpretation is plausible because, in children, the vocal tract is shorter and the larynx has not yet fully descended, which is associated with higher F1 (Vorperian et al, 2005; Yoon and Ha, 2022). From this perspective, the F1 increase in *aegyo* may reflect an attempt to stylize children's vocal-tract configuration, especially along the pharyngeal dimension, which is easier to manipulate than oral-cavity length.

The findings show both similarities and differences from previous studies of children's corner-vowel space. Like Pettinato et al. (2016), *aegyo* exhibits a larger corner-vowel space. In contrast, Chung et al. (2012) report that young children produce smaller corner-vowel spaces than adults when vowel space is calculated with normalization of the vocal tract length. However, whereas Pettinato et al. (2016) reported vowel space expansion arising from coordinated changes in both F1 and F2, the present results show that the expansion is driven primarily by increases in F1, with more limited contributions from F2. This suggests that adults may attempt to iconically reflect the shorter vocal tract of children, however, their ability to modify the oral cavity (as reflected in F2) is limited.

Finally, our analysis is limited to acoustic measurements and therefore does not allow us to determine whether the observed pattern is driven primarily by laryngeal raising, tongue or lip position adjustments, or a combination of all. For this reason, further articulatory study is needed to test whether this F1 increase reflects pharyngeal shortening and/or other articulatory adjustments.

5. **Conclusion**

Korean *aegyo* speech is characterized by a systematic increase in F1 values, reflecting global lowering across the vowel system. As a result of this F1-dominated shift, the overall vowel space is larger in *aegyo* than in non-*aegyo* speech. Selective fronting is observed only for front vowels. The disappearance of the F2 centroid effect when the full vowel inventory is analyzed indicates that the apparent fronting is limited to a subset of vowels rather than the entire system.

More broadly, our findings indicate that *aegyo* is characterized primarily through F1 increase. While this pattern may reflect articulatory adjustments associated with iconically imitating the shorter vocal tract of children, further articulatory investigation is required to determine the extent to which this F1-dominated pattern reflects pharyngeal shortening, tongue or lip position adjustment, or all.

**Author Declarations**

*Conflict of Interest*

The authors have no conflicts of interest to disclose.

*Ethics Approval*

The recordings were collected under the institutional regulations in place at the time and were determined to be exempt from formal ethics review. All participants provided informed consent prior to the recordings.

**References (BIBLIOGRAPHIC STYLE)**